\documentclass[conference]{IEEEtran}
\IEEEoverridecommandlockouts
\usepackage{amsmath,amssymb,amsfonts}
\usepackage{algorithmic}
\usepackage{graphicx}
\usepackage{caption} 
\usepackage{textcomp}
\usepackage{xcolor}
\usepackage{bm}
\usepackage{booktabs}
\usepackage[backend=bibtex]{biblatex}  
\newcommand{\W}{\bm{W}}
\newcommand{\B}{\bm{B}}
\newcommand{\A}{\bm{A}}

\usepackage{hyperref}
\usepackage{stmaryrd}
\usepackage{amsmath, amssymb}
\usepackage{enumitem}
\usepackage{csquotes}
\usepackage{amsthm}

\def\BibTeX{{\rm B\kern-.05em{\sc i\kern-.025em b}\kern-.08em
    T\kern-.1667em\lower.7ex\hbox{E}\kern-.125emX}}

\theoremstyle{plain}
\newtheorem{theorem}{Theorem}

\theoremstyle{definition}
\newtheorem{definition}{Definition}
\newtheorem{assumption}{Assumption}
\theoremstyle{remark}

\begin{document}

\title{Localized LoRA: A Structured Low-Rank Approximation for Efficient Fine-Tuning\\

\author{\IEEEauthorblockN{Babak Barazandeh}
\IEEEauthorblockA{\textit{AI Risk and Vulnerability Alliance} \\
babakbarazandeh@gmail.com}
\\[1em]
\hfill
\IEEEauthorblockN{Om Rajyaguru}
\IEEEauthorblockA{\textit{AI Risk and Vulnerability Alliance} \\
om.rajyaguru@outlook.com}
\and
\IEEEauthorblockN{Subhabrata Majumdar}
\IEEEauthorblockA{\textit{Vijil} \\
subho@vijil.ai}
\\[1em]
\hfill
\IEEEauthorblockN{George Michailidis}
\IEEEauthorblockA{\textit{University of California, Los Angeles} \\
gmichail@stat.ucla.edu}
}
}

\maketitle

\begin{abstract}
Parameter-efficient fine-tuning (PEFT) methods, such as LoRA, offer compact and effective alternatives to full model fine-tuning by introducing low-rank updates to pretrained weights. However, most existing approaches rely on global low-rank structures, which can overlook spatial patterns spread across the parameter space. In this work, we propose \textbf{Localized LoRA}, a generalized framework that models weight updates as a composition of low-rank matrices applied to structured blocks of the weight matrix. This formulation enables dense, localized updates throughout the parameter space—without increasing the total number of trainable parameters. We provide a formal comparison between global, diagonal-local, and fully localized low-rank approximations, and show that our method consistently achieves lower approximation error under matched parameter budgets. Experiments on both synthetic and practical settings demonstrate that Localized LoRA offers a more expressive and adaptable alternative to existing methods, enabling efficient fine-tuning with improved performance.
\end{abstract}

\begin{IEEEkeywords}
Parameter-efficient fine-tuning, Low-rank adaptation, LoRA, Matrix approximation, Local low-rank structure, Large language models, Structured sparsity, Neural network compression
\end{IEEEkeywords}

\section{Introduction}
Modern natural language processing (NLP) is increasingly driven by Large Language Models (LLMs), which exhibit strong generalization across diverse tasks. While full fine-tuning remains the most effective adaptation strategy in theory, the immense scale of LLMs renders this approach computationally prohibitive for most practical scenarios. The combination of large model sizes, task diversity, and limited compute budgets necessitates more efficient tuning strategies.

Parameter-Efficient Fine-Tuning (PEFT) methods address this challenge by adapting only a small subset of parameters. Notable among these is Low-Rank Adaptation (LoRA)~\cite{hu2022lora}, which introduces trainable low-rank matrices without modifying the base model. Other strategies such as adapters~\cite{houlsby2019parameter}, prefix-tuning~\cite{li2021prefix},  and modular approaches like MoRAL~\cite{yang2024moral} (which combines mixture-of-experts with LoRA to support efficient lifelong learning in large language models) enable robust knowledge retention and improved adaptation to new tasks via question-answer based fine-tuning.

Recent efforts include Bi-Share LoRA~\cite{bishare2025}, which combines intra-layer and inter-layer parameter sharing to improve memory efficiency across tasks, and XGBLoRA~\cite{zhang2024less}, which applies boosting-inspired iterative updates for performance gains using rank-1 matrices.

Further refinements target specific limitations of LoRA. LoR2C~\cite{zhao2025lor2c} mitigates vanishing gradients via residual low-rank paths, while GoRA~\cite{he2025gora} leverages gradient sensitivity to adaptively select ranks. LoLDU~\cite{shi2024lold} introduces structured lower-diagonal-upper decomposition for efficient and orthogonal updates, and SVFit~\cite{sun2024svfit} improves initialization by leveraging top singular values from the pretrained model. DeLoRA~\cite{bini2025delora} separates adaptation angle from strength, boosting robustness across tasks.

Recent work has expanded PEFT into non-Euclidean spaces. HypLoRA~\cite{yang2024hyperbolic} explores hyperbolic geometry to better represent the hierarchical structure of tokens and embeddings. Survey efforts such as~\cite{yang2025lowrank} have comprehensively reviewed LoRA extensions across domains.

Other approaches tackle efficiency from structural or compositional angles. MoKA~\cite{yu2025moka} combines Kronecker products with a mixture-of-experts (MoE) routing, and K-LoRA~\cite{ouyang2025klora} enables training-free fusion of content and style LoRAs by comparing top-K elements across matrices. Knowledge preservation is another concern; LoRAMoE~\cite{dou2025loramoe} uses MoE-style routers to integrate multiple LoRA adapters while maintaining world knowledge, and KELE~\cite{zhang2024enhancing} introduces knowledge erasure to improve multi-hop reasoning after edits.

Multimodal models add further complexity. Tuning vision-language models faces issues like task specialization and catastrophic forgetting. A taxonomy of such challenges and tuning strategies is provided by~\cite{huang2025keeping}, covering selective tuning, additive adaptation, and reparameterization. Evaluation initiatives like INSTRUCTEVAL~\cite{chia2023instructeval} aim to standardize and benchmark instruction-tuned LLMs across alignment, reasoning, and writing.

Alternative PEFT strategies enhance LoRA’s practicality by improving scalability, modularity, and efficiency. Approaches such as VeRA~\cite{kopiczko2023vera} reduce storage costs through sharing low-rank matrices across layers and tuning small scaling vectors per layer, while others explore dynamic rank allocation~\cite{mao2024dora}, vector bank-based tuning~\cite{liu2024vb}, and cross-task generalization using composable LoRA modules~\cite{huang2023lorahub}.
These developments position LoRA as a flexible framework for efficient and adaptive learning across diverse tasks.

This paper revisits PEFT from a matrix approximation perspective, specifically building on recent advances in localized low-rank adaptation. While standard LoRA assumes a global low-rank structure across entire weight matrices, recent work like MELoRA~\cite{ren2024MELoRA} has demonstrated the benefits of applying low-rank updates to localized diagonal regions of the weight matrix. However, MELoRA's restriction to diagonal blocks, while effective, represents only a special case of a broader class of spatially-aware adaptation strategies.

To address this limitation, we propose \textbf{Localized LoRA}, a principled framework of performing localized low-rank adaptation on weight matrices with arbitrary spatial structures. Rather than constraining updates to diagonal regions, our framework enables low-rank adaptation across the full spatial extent of the parameter space through structured block partitioning. This generalization maintains the parameter efficiency benefits of MELoRA while significantly expanding its representational capacity.

Our key contributions are:
\begin{enumerate}[leftmargin=*]
    \item We formalize a general framework for spatially structured low-rank adaptation, 
    \item Compared to restricted methods such as MELoRA, We demonstrate that this generalization achieves superior approximation quality under matched parameter budgets, and 
    \item  We show consistently good performance in both synthetic reconstruction tasks and practical fine-tuning scenarios.
\end{enumerate}

\section{Methodology}

In this section, we present our proposed method, which introduces a more flexible local low-rank structure for fine-tuning large models. We begin by establishing a general framework that unifies existing approaches to localized low-rank adaptation. To that end, it decomposes the a fine-tuning weight update $\Delta \mathbf{W}$ as a composition of low-rank matrices applied to structured spatial regions.

\subsection{Preliminaries}

We begin by formalizing a few preliminary concepts.

\begin{definition}[Globally Low-Rank Matrix]\label{def:globally-low-rank}
A matrix \( \mathbf{M} \in \mathbb{R}^{m \times n} \) is said to be \emph{globally low-rank} if its rank is significantly smaller than the minimum of its dimensions:
\[
\mathrm{rank}(\mathbf{M}) = r \ll \min(m, n).
\]
This implies that the entire matrix approximately lies in a low-dimensional linear subspace.
\end{definition}

\begin{definition}[Locally Low-Rank Matrix]\label{def:locally-low-rank}
A matrix \( \mathbf{M} \in \mathbb{R}^{m \times n} \) is said to be \emph{locally low-rank} if it can be partitioned into a collection of submatrices \( \{\mathbf{M}_k\} \), where each \( \mathbf{M}_k \in \mathbb{R}^{m_k \times n_k} \) satisfies:
\[
\mathrm{rank}(\mathbf{M}_k) = r_k \ll \min(m_k, n_k), \quad \forall k.
\]
In this case, the full matrix \( \mathbf{M} \) may be high-rank overall, but exhibits low-rank structure within each localized region.
\end{definition}

\paragraph*{Low-rank Adaptation (LoRA)}
One of the most commonly used techniques for parameter-efficient fine-tuning of Large Language Models (LLMs) is LoRA. LoRA assumes that \( \Delta \W \), the weight update done during the fine-tuning of an LLM, can be approximated by the product of two low-rank matrices:
\begin{equation}
  \W^* = \W + \Delta \W = \W + \B \A,
\end{equation}
where \( \B \in \mathbb{R}^{d \times r} \), \( \A \in \mathbb{R}^{r \times d} \), and \( \W \in \mathbb{R}^{d \times d} \) is the weight matrix of the pretrained LLM. 

From the perspective of Definition~\ref{def:globally-low-rank}, this formulation approximates the weight update as a \emph{globally low-rank} matrix. While this approach is highly parameter-efficient, it may limit expressiveness, particularly when the desired adaptation requires capturing diverse or region-specific variations. This limitation arises because global low-rank constraints force all spatial regions of the weight matrix to share the same low-dimensional subspace, preventing specialized adaptations that may be needed for different feature groups or semantic components~\cite{kopiczko2023vera,ren2024MELoRA}. Recent empirical evidence suggests that effective fine-tuning often requires heterogeneous updates across different parts of the parameter space, as different layers and dimensions in large models tend to specialize for distinct representational functions~\cite{he2025gora,huang2023lorahub}.

This raises an important question:

\begin{displayquote}
\emph{
Can we design a new formulation that models the weight update as \textbf{locally low-rank}, without significantly increasing the number of trainable parameters?
}
\end{displayquote}

Several recent works explore alternative parameterizations of the weight update. The most relevant to our work is MELoRA~\cite{ren2024MELoRA}, which we generalize over.

\subsection{Localized LoRA}

To capture structured variations in weight matrix updates, we propose a blockwise low-rank adaptation strategy that applies localized updates to different regions of the weight matrix. Specifically, we partition the weight matrix \( \W \in \mathbb{R}^{d \times d} \) into equal blocks of size \( K \times K \) by dividing its rows and columns into \( K \) segments. Each block \( (i, j) \) is assigned an independent low-rank adapter, parameterized by matrices \( \A_{ij} \in \mathbb{R}^{r_{ij} \times d/K} \) and \( \B_{ij} \in \mathbb{R}^{d/K  \times r_{ij}} \). The full adaptation is constructed by assembling these blockwise updates into a structured matrix:

\begin{equation}
\mathcal{B} \llbracket \{ \B_{ij}, \A_{ij} \}_{i,j=1}^K \rrbracket =
\begin{bmatrix}
\B_{11} \A_{11} & \cdots & \B_{1K} \A_{1K} \\
\vdots & \ddots & \vdots \\
\B_{K1} \A_{K1} & \cdots & \B_{KK} \A_{KK}
\end{bmatrix}.
\end{equation}

This operator \( \mathcal{B} \llbracket \cdot \rrbracket \) returns a \( d \times d \) matrix in which each \( (i,j) \)-th block is filled with the product \( \B_{ij} \A_{ij} \), a low-rank update specific to that block. The adapted weight matrix is then given by:

\begin{equation}
\W^* = \W + \mathcal{B} \llbracket \{ \B_{ij}, \A_{ij} \}_{i,j=1}^K \rrbracket.
\end{equation}

Although in general each block \( (i,j) \) may have its own rank \( r_{ij} \), for clarity and consistency we assume a uniform rank \(  r_{\text{block}} \) across all blocks here on. This simplifies analysis and implementation without loss of generality. As we shall see later, this required rank \( r_{\text{block}} \) in localized LoRA is typically much lower than the global rank used in standard LoRA, as each block captures a more spatially focused component of the adaptation. This localized structure enables our method to achieve high expressiveness with significantly fewer trainable parameters.

\subsection{Comparison with Existing Methods}
By assembling the full update as a sum of these blockwise low-rank components, our formulation supports fine-grained, spatially aware adaptation across the entire weight matrix. This improves representational flexibility while preserving the efficiency advantages of low-rank parameterization. Notably, our approach recovers the standard LoRA as a special case when \( K = 1 \), and can thus be viewed as \emph{locally low-rank generalization of LoRA}, balancing the efficiency of the parameters with the enhanced modeling capacity.

As a formal comparison, standard LoRA applies an identity operator, while MELoRA~\cite{ren2024MELoRA} applies the diagonal-block operator $\mathcal{D} \llbracket \cdot \rrbracket$ while doing $\Delta \W$ updates:

\begin{equation}
\mathcal{D} \llbracket \{ \mathbf{B}_i, \mathbf{A}_i \}_{i=1}^N \rrbracket =
\begin{bmatrix}
\mathbf{B}_1 \mathbf{A}_1 & 0 & \cdots & 0 \\
0 & \mathbf{B}_2 \mathbf{A}_2 & \cdots & 0 \\
\vdots & \vdots & \ddots & \vdots \\
0 & 0 & \cdots & \mathbf{B}_N \mathbf{A}_N
\end{bmatrix},
\end{equation}

where $\mathbf{B}_i \in \mathbb{R}^{d/N \times r_{\text{diag}}}$ and $\mathbf{A}_i \in \mathbb{R}^{r_{\text{diag}} \times d/N}$. This is a special case under our proposal, where we constrain $\mathbf{B}_{ij} \mathbf{A}_{ij} = 0$ for all $i \neq j$. This design captures within-block dependencies but inherently cannot model cross-block interactions because of its zero off-diagonal structure. We empirically demonstrate later (Section~\ref{subsec:mnist}) that this diagonal constraint can substantially restrict the representational capacity of the update matrix compared to more flexible local low-rank formulations.

\section{Approximation Capacity and Trainable Parameters}

The fundamental trade-off PEFT addresses in neural network adaptation is that of achieving high task-specific expressivity in weight updates while minimizing the number of trainable parameters. While Supervised Fine-Tuning (SFT, or `full' fine-tuning) is highly expressive and can enable high task or domain adaptation, its number of trainable parameters is the same as the size of the neural network. Standard LoRA constrains the update matrix $\Delta \mathbf{W}$ to have a global low-rank structure, allowing compact parameterization through the factorization $\Delta \mathbf{W} = \mathbf{B}\mathbf{A}$. However, this global constraint may be overly restrictive when the optimal adaptation exhibits spatially heterogeneous structure---different regions of the weight matrix requiring different types of low-rank updates that cannot be efficiently captured by a single global factorization.

Localized LoRA addresses this limitation by enabling spatially-aware low-rank decomposition while maintaining parameter efficiency. By distributing the representational capacity across structured spatial regions, we can achieve approximation quality similar to global methods while using a much smaller number of trainable parameters.

\subsection{Theoretical Analysis}

We adapt the theoretical framework from \cite{zeng2024the} to analyze when Localized LoRA can efficiently adapt to targets with spatial low-rank structure. Formally, consider the situation where we would like to take an arbitrary frozen Fully Connected Neural Network (FNN), say $f$, and adapt it to represent a simpler FNN $f^*$, both operating on input sample space $\mathcal X$.

\begin{assumption}[Spatially Localized Target]
\label{ass:spatial-target}
The target weight update $\boldsymbol{\Delta}_{\text{target}} \in \mathbb{R}^{d \times d}$ can be partitioned into $K \times K$ spatial blocks such that each block $\boldsymbol{\Delta}_{ij} \in \mathbb{R}^{(d/K) \times (d/K)}$ satisfies $\text{rank}(\boldsymbol{\Delta}_{ij}) \leq r_{\text{local}}$ where $r_{\text{local}} \ll d/K$.
\end{assumption}

\begin{theorem}[Exact Spatial Adaptation]
\label{thm:spatial-exact}
Let $f_0$ be a frozen FNN and $f^*$ be a target FNN whose weight differences satisfy Assumption~\ref{ass:spatial-target}. Under the non-singularity conditions of \cite{zeng2024the}, there exist Localized LoRA adapters $\{\mathbf{B}_{ij}, \mathbf{A}_{ij}\}_{i,j=1}^K$ such that the adapted model $\tilde{f}$ exactly represents the target model $f^*$, i.e., $\tilde{f}(x) = f^*(x)$ for all $x \in \mathcal{X}$.
\end{theorem}

\begin{proof}[Proof Sketch]
We apply Theorem 3 from \cite{zeng2024the} to each spatial block independently. Defining $r_{\text{local}} = \max_{i,j} rank (\boldsymbol{\Delta}_{ij})$, their result guarantees that rank-$r_{\text{block}}$ LoRA adapters can exactly represent each block when $r_{\text{block}} \geq r_{\text{local}}$. The global reconstruction follows by assembling these block-wise exact representations using our spatial operator $\mathcal{B}[\cdot]$.
\end{proof}

\begin{theorem}[Spatial Approximation Error]
\label{thm:spatial-approx}
Under the same setting as Theorem~\ref{thm:spatial-exact}, when $r_{\text{block}} < \max_{i,j} \text{rank}(\boldsymbol{\Delta}_{ij})$, there exist Localized LoRA adapters such that the approximation error satisfies:
$$\mathbb{E}[\|f^*(x) - \tilde{f}(x)\|^2] \leq \beta \sum_{i,j=1}^K \sigma_{r_{\text{block}}+1}^2(\boldsymbol{\Delta}_{ij}),$$
where $\beta$ depends on the network parameters and input distribution (as defined in Theorem 5 of \cite{zeng2024the}), and $\sigma_{r_{\text{block}}+1}(\boldsymbol{\Delta}_{ij})$ is the $(r_{\text{block}}+1)$-th largest singular value of block $\boldsymbol{\Delta}_{ij}$.

\end{theorem}

\begin{proof}[Proof Sketch]
We apply Theorem 5 from \cite{zeng2024the} to each block. Their approximation bound gives us the error for each block as $\sigma_{r_{\text{block}}+1}^2(\boldsymbol{\Delta}_{ij})$. The global error is the sum of blockwise errors due to the orthogonality of our spatial decomposition.
\end{proof}

Thus, when the target has favorable spatial structure (Assumption~\ref{ass:spatial-target}), Localized LoRA can achieve the same approximation quality as standard LoRA while using fewer parameters.

\subsection{Comparison of Trainable Parameters}
A primary motivation for our approach is to maximize expressive power while minimizing the number of trainable parameters. In this section, we compare the parameter complexity of three methods: standard LoRA, MELoRA, and our proposed Localized LoRA. For consistency with earlier sections, we use \( r \) to denote the global rank in LoRA, \( r_{\text{diag}} \) for the rank used in each diagonal block of MELoRA, and \( r_{\text{block}} \) for the rank assigned to each local block in our method.

Although these ranks share similar notation, they differ significantly in both meaning and practical scale. As established in previous sections and further substantiated by our experiments, Localized LoRA achieves strong performance using a much smaller \( r_{\text{block}} \) than the global rank \( r \) typically required by LoRA. This efficiency arises from our method's ability to spatially distribute low-rank adaptations across the matrix, reducing the representational load on any individual block. While MELoRA also benefits from localized structure—allowing for a reduced diagonal rank \( r_{\text{diag}} \)—its restriction to diagonal-only regions limits its flexibility. In contrast, our approach generalizes this idea by enabling low-rank updates across the entire matrix, including off-diagonal interactions, resulting in greater adaptability with fewer parameters.

\textbf{LoRA:}  
In standard LoRA, the update to a weight matrix \( \W \in \mathbb{R}^{d \times d} \) is approximated by a global rank-\( r \) matrix \( \Delta \W = \B \A \), where \( \B \in \mathbb{R}^{d \times r} \) and \( \A \in \mathbb{R}^{r \times d} \). The number of trainable parameters is:
\[
\#\text{params}_{\text{LoRA}} = 2dr.
\]

\textbf{MELoRA:}  
This approach divides the weight matrix into \( N \) diagonal blocks. Each block is adapted with a rank-\( r_{\text{diag}} \) update using \( \B_i \in \mathbb{R}^{d/N \times r_{\text{diag}}} \) and \( \A_i \in \mathbb{R}^{r_{\text{diag}} \times d/N} \), assuming all blocks use the same rank. The total parameter count becomes:
\[
\#\text{params}_{\text{MELoRA}} = N \cdot 2 \cdot \frac{d}{N} \cdot r_{\text{diag}} = 2dr_{\text{diag}}.
\]
If we set \( r_{\text{diag}} = \frac{r}{N} \), this becomes:
\[
\#\text{params}_{\text{MELoRA}} = \frac{2dr}{N},
\]
which is smaller than the parameter count of standard LoRA.

\textbf{Localized LoRA:}  
Our method partitions the matrix into \( K^2 \) spatial blocks, not restricted to diagonals. Each block uses a rank-\( r_{\text{block}} \) update, requiring:
\[
\#\text{params}_{\text{Localized}} = K^2 \cdot 2 \cdot \frac{d}{K} \cdot r_{\text{block}} = 2dr_{\text{block}}K.
\]
Setting \( K = \sqrt{N} \) and choosing \( r_{\text{block}} = \frac{r}{N\sqrt{N}} \), we obtain:
\[
\#\text{params}_{\text{Localized}} = 2d \cdot \frac{r}{N\sqrt{N}} \cdot \sqrt{N} = \frac{2dr}{N},
\]
matching the parameter count of MELoRA. This demonstrates that Localized LoRA retains the parameter efficiency of MELoRA while enabling a more expressive, fully localized low-rank structure.

\textbf{Example:}  
Let \( d = 256 \), \( r = 16 \), and \( N = 4 \). Then \( \sqrt{N} = 2 \), so \( K = 2 \), and:
\begin{itemize}
  \item \textbf{LoRA:}
  \[
  \#\text{params}_{\text{LoRA}} = 2 \cdot 256 \cdot 16 = 8192
  \]
  
  \item \textbf{MELoRA}, with \( r_{\text{diag}} = \frac{16}{4} = 4 \):
  \[
  \#\text{params}_{\text{MELoRA}} = 2 \cdot 256 \cdot 4 = 2048
  \]
  
  \item \textbf{Localized LoRA}, with \( r_{\text{block}} = \frac{16}{4 \cdot 2} = 2 \), and \( K = 2 \):
  \[
  \#\text{params}_{\text{Localized}} = 2 \cdot 256 \cdot 2 \cdot 2 = 2048
  \]
\end{itemize}

Localized LoRA uses only a quarter of the parameters compared to LoRA and matches MELoRA in parameter count. Yet, it achieves superior results in both matrix reconstruction and downstream adaptation tasks. This gain arises from its ability to model both local and non-diagonal interactions through spatially distributed low-rank decomposition, enabling greater expressiveness without additional cost.

\section{Numerical Experiments}

In this section, we present two experiments designed to evaluate the effectiveness of spatially structured low-rank adaptation methods. The first is a controlled matrix approximation task on MNIST digits, where LoRA, MELoRA, and Localized LoRA are constrained to use identical parameter budgets, allowing us to isolate the impact of spatial decomposition on approximation quality. The second is a realistic fine-tuning scenario, where we adapt a frozen MLP (Multi-Layer Perceptron) pretrained on digits 0--4 to classify digits 5--9. In both experiments, we assess performance under tightly controlled conditions to highlight the advantages of spatial flexibility in low-rank updates. Results are measured in terms of reconstruction error and classification accuracy, respectively.

\subsection{Experiment 1: Binary Classification}
\label{subsec:mnist}

To further evaluate the capacity of our method to model spatially structured low-rank information, we conduct an experiment on the MNIST dataset. We treat each digit image as a \( 28 \times 28 \) matrix and approximate it using low-rank decompositions with tightly matched parameter budgets. This setting serves as a clean and interpretable benchmark for evaluating local versus global low-rank approximations.

\paragraph{Setup}
We select the first test image of the digit ``2'' (randomly chosen) from MNIST, which has an effective matrix rank of 15. Each method (LoRA, MELoRA, Localized LoRA) is constrained to use exactly \( 2rd = 224 \) parameters, where \( d = 28 \) and \( r=4 \) is the baseline LoRA rank. The methods are configured as follows:

\begin{itemize}
\item \textbf{LoRA}: Applies a global rank-$r = 4$ approximation to the full image.
\item \textbf{MELoRA}: Divides the image into $N = 4$ diagonal blocks, each approximated with rank-$r_{\text{diag}} = 4$.
\item \textbf{Localized LoRA}: Partitions the image into $K = 2$ spatial regions along each axis (producing 4 blocks total), with each block approximated using rank-$r_{\text{block}} = 2$.
\end{itemize}

\noindent All methods are implemented using truncated SVD. The reconstruction quality is measured using normalized Frobenius error:
\[
\text{Error} = \frac{\|W_{\text{true}} - W_{\text{approx}}\|_F}{\|W_{\text{true}}\|_F}.
\]

\paragraph{Results}
Table~\ref{tab:mnist-matched-params} reports the reconstruction errors and effective parameter counts for each method.

\begin{table}[h]
\centering
\small
\setlength{\tabcolsep}{6pt}
\begin{tabular}{lccc}
\toprule
\textbf{Method} & \textbf{Rank} & \textbf{\#Params} & \textbf{Error} \\
\midrule
LoRA (Global)   & 4 (global) & 224 & 0.2313 \\
MELoRA          & 4 (diag)   & 224 & 0.9071 \\
Localized LoRA (Ours)& 2 (block)   & 224 & \textbf{0.2119} \\
\bottomrule
\end{tabular}
\caption{Normalized reconstruction error with matched parameter budget ($2rd = 224$). Localized LoRA achieves the best approximation.}

\label{tab:mnist-matched-params}
\end{table}

\paragraph{Discussion}
Despite using the same number of trainable parameters, our proposed Localized LoRA method achieves the lowest reconstruction error. While MELoRA is limited to diagonal structure and thus fails to recover off-diagonal stroke information, Localized LoRA distributes its modeling capacity across the spatial layout of the image. Even LoRA, though globally applied, underperforms compared to Localized LoRA in this constrained setting.

\begin{figure*}[t]
    \centering
    \includegraphics[width=\textwidth]{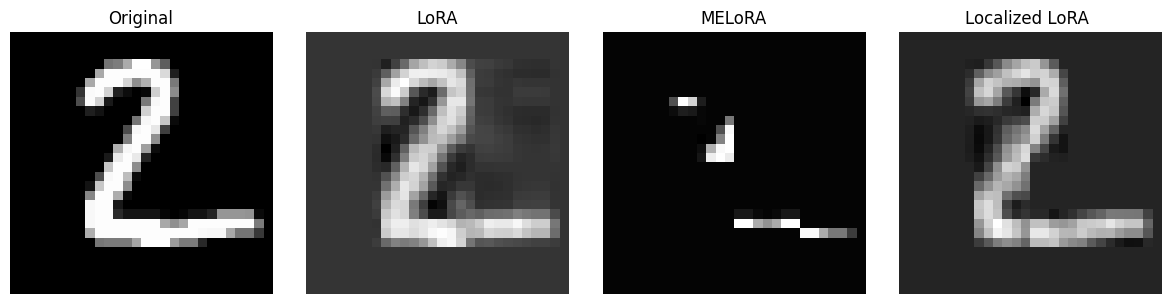} 
    \caption{Reconstruction of an MNIST digit using different low-rank approximation methods. From left to right: original image, LoRA, MELoRA, and our proposed Localized LoRA. Localized LoRA achieves better fidelity in recovering both local and global stroke structures under the same parameter budget.}
    \label{fig:matrix_comparison}
\end{figure*}

Figure~\ref{fig:matrix_comparison} visualizes the reconstructions produced by each method. The qualitative differences align with the quantitative findings: Localized LoRA preserves both the digit’s global shape and local stroke details, whereas MELoRA introduces heavy distortion, and LoRA fails to capture fine structure.

This result demonstrates that leveraging localized spatial structure—while maintaining parameter efficiency—can yield significant improvements in approximation quality. This supports the broader claim that general-purpose low-rank updates benefit from spatial flexibility, particularly in image-like or structured domains.

\subsection{Experiment 2: Multi-class Classification}

To evaluate the effectiveness of parameter-efficient adaptation methods, we perform a domain adaptation experiment on MNIST. Specifically, we adapt a model pretrained on digits 0--4 to perform well on digits 5--9 using only a limited number of trainable parameters. While full fine-tuning of the entire model serves as a strong baseline, it involves updating all parameters—resulting in significantly higher memory usage and storage costs. In contrast, we compare several low-rank adaptation techniques that aim to match the performance of full fine-tuning with a fraction of the trainable parameters.

\paragraph{Setup}
We first train a feedforward MLP model on digits 0--4. This model consists of a single hidden layer with ReLU activation, followed by a classification layer. During adaptation, both layers are frozen, and trainable low-rank adapters are inserted before the activation, modifying the hidden representation learned by the frozen network.

We evaluate the following parameter-efficient adaptation methods:

\begin{itemize}
\item \textbf{LoRA}: A global low-rank adapter whose output is added to the frozen input-to-hidden transformation.
\item \textbf{MELoRA}: The input is divided into $N$ blocks, each mapped to a corresponding slice of the hidden representation. A separate low-rank adapter is applied independently to each block, and their outputs are additively combined with the base model’s output.
\item \textbf{Localized LoRA}: The input is reshaped into a $K \times K$ spatial grid. Each grid cell is processed by a local low-rank adapter, and their outputs are added to the frozen transformation.
\end{itemize}

We also compare against full fine-tuning, where all parameters of the network are updated. In contrast, all adaptation methods considered here modify a shared frozen model by adding trainable low-rank adapters to the linear layers:
\begin{table}[h]
\centering
\begin{tabular}{@{}ll@{}}
\toprule
\textbf{Component}        & \textbf{Details} \\
\midrule
Input                     & \parbox[t]{5cm}{Flattened MNIST image \\(\(28 \times 28 = 784\))} \\
Linear Layer (Frozen)     & \texttt{Linear(784, 64)} \\
Low-Rank Adapter (Trainable) & \parbox[t]{5cm}{LoRA, MELoRA, or Localized LoRA} \\
Activation                & \texttt{ReLU} \\
Linear Classifier (Frozen)& \texttt{Linear(64, 5)} \\
Output                    & \texttt{LogSoftmax} \\
\bottomrule
\end{tabular}
\caption{Frozen feedforward architecture. Low-rank adapters are added before the activation function. Only the adapter parameters are trainable.}
\label{tab:frozen-mlp-architecture}
\end{table}


\paragraph{Hyperparameter Sweep}
Each method is fine-tuned under multiple configurations. We sweep learning rates in $\{1\mathrm{e}{-4}, 2\mathrm{e}{-4}, 5\mathrm{e}{-4}\}$ and report the best-performing configuration based on validation loss:
\begin{itemize}
    \item \textbf{LoRA}: Ranks $r \in \{2, 4, 8, 16, 32\}$.
    \item \textbf{MELoRA}: Ranks $r_{\text{diag}} \in \{2, 4, 8, 16, 32\}$ with block counts $N \in \{2, 4, 8, 16\}$.
    \item \textbf{Localized LoRA}: Ranks $r_{\text{block}} \in \{2, 4, 8, 16\}$ with grid sizes $K \in \{2, 4, 8, 16\}$.
\end{itemize}

Note that for \textit{MELoRA} and \textit{Localized LoRA}, the reported rank refers to the internal rank of each small adapter block. Specifically, MELoRA uses $N$ such blocks and Localized LoRA applies $K^2$ adapters. Configurations where the block rank approached or exceeded block size were excluded to avoid overparameterization. For clarity and fairness, we report only the best-performing configuration at each parameter budget (i.e., total number of trainable parameters), rather than comparing methods using the same rank or block count. This allows us to evaluate each method’s efficiency in terms of accuracy per parameter.

\begin{figure}[h]
\centering
\includegraphics[width=0.9\linewidth]{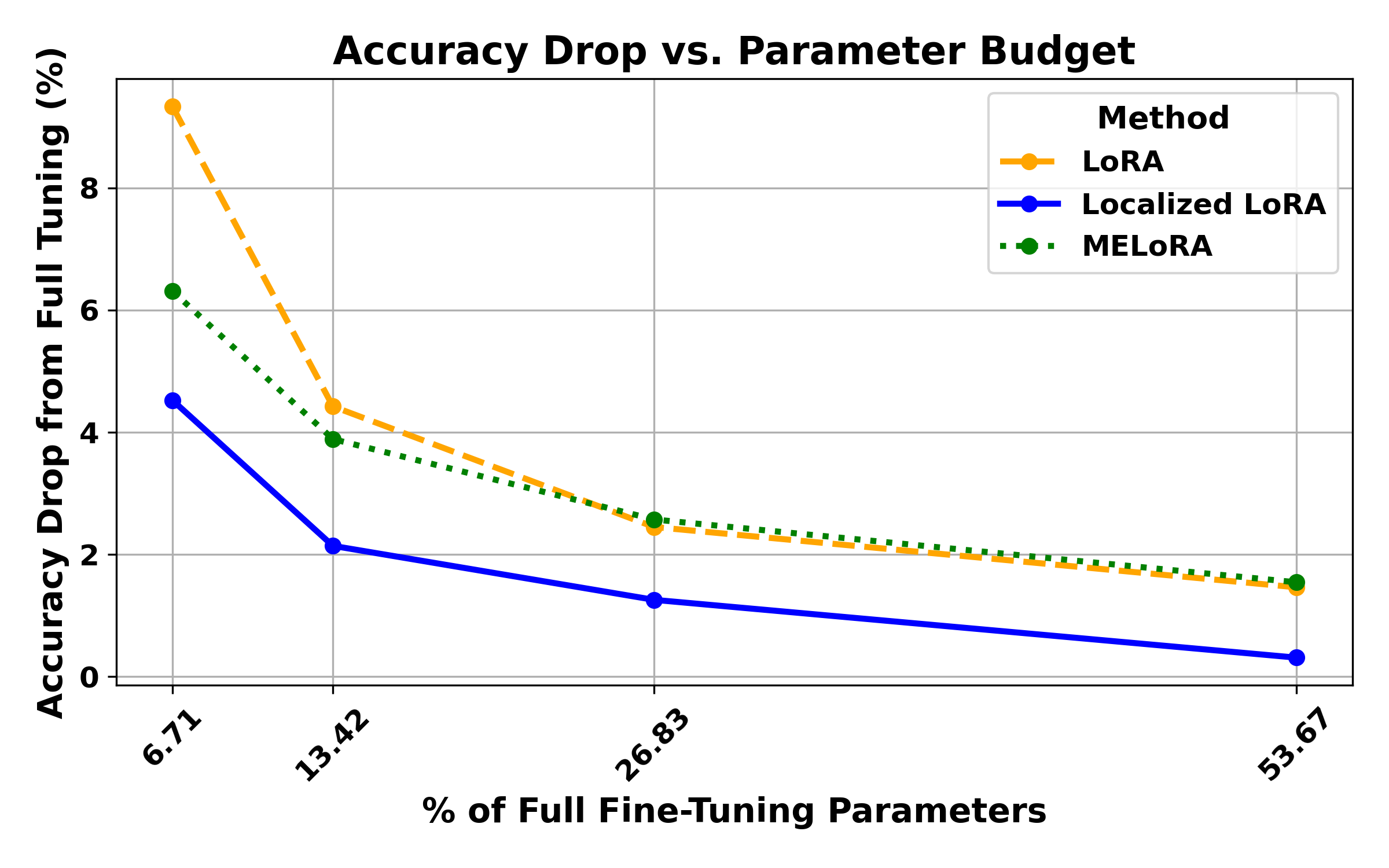}
\caption{Accuracy drop (\%) compared to SFT as a function of parameter budget. SFT uses 50,565 parameters and achieves 97.6\% accuracy.}
\label{fig:accuracy-drop}
\end{figure}

\paragraph{Results}
Figure~\ref{fig:accuracy-drop} shows the accuracy drop of each method compared to full fine-tuning, plotted against the proportion of trainable parameters used. Localized LoRA consistently achieves lower accuracy drop across the budget range, outperforming both LoRA and MELoRA, especially in the low-parameter regime. Notably, Localized LoRA matches the performance of MELoRA while using nearly half as many parameters, and exceeds LoRA even when LoRA is granted a larger parameter budget. These results highlight the strong efficiency-accuracy trade-off offered by spatially-aware low-rank adaptation.

\section{Conclusion}
Our experiments highlight the effectiveness of structured low-rank adaptation methods for domain adaptation under tight parameter budgets. By introducing small trainable adapters into a shared frozen model, structured low-rank methods can match or approach full fine-tuning performance using only a fraction of the trainable parameters. Among these, Localized LoRA consistently shows the most favorable trade-off between accuracy and parameter count, outperforming LoRA and MELoRA especially in low-budget regimes. These results suggest that spatial decomposition, as introduced in Localized LoRA, enables more efficient and scalable fine-tuning, making it well suited for resource-constrained adaptation tasks.

Beyond computational efficiency, more expressive PEFT methods like Localized LoRA can significantly enhance AI safety evaluation techniques. By enabling more precise and localized model adaptations with fewer parameters, these methods facilitate better alignment fine-tuning procedures that can more effectively steer model behavior without compromising the underlying capabilities. The improved parameter efficiency allows for more comprehensive safety evaluations through rapid iteration over multiple adaptation strategies, enabling researchers to systematically test model robustness across diverse scenarios while maintaining computational feasibility for safety-critical applications.

\section*{Acknowledgments}
We are grateful to the anonymous reviewers for their helpful feedback. GM is supported by United States National Science Foundation (NSF) grants 2319593 and 2348640.

\printbibliography

\end{document}